\documentclass[runningheads]{llncs}
\usepackage[T1]{fontenc}
\usepackage{graphicx}
\usepackage{multirow}
\usepackage{tabularx} 
\usepackage{array}
\usepackage{booktabs} 
\usepackage{subfigure}
\usepackage{threeparttable}
\usepackage{siunitx}

\begin{document}
\title{Boosting ECG Classification Performance by Pre-training with Synthesized Data}
\titlerunning{Boosting ECG Classification with Synthesized Data}
%
\author{
Naoki Nonaka \and
Jun Seita
}
%
\authorrunning{Nonaka et al.}
%
\institute{Advanced Data Science Project, \\RIKEN Information R\&D and Strategy Headquarters}
%
\maketitle              

\newcolumntype{C}{>{\centering\arraybackslash}p{2.5cm}} 
\newcolumntype{D}{>{\centering\arraybackslash}p{1.5cm}} 
\newcolumntype{E}{>{\centering\arraybackslash}p{4.7cm}} 
\newcolumntype{R}{>{\raggedleft\arraybackslash}X} 

\begin{abstract}

Deep Neural Networks (DNNs) typically require extensive datasets for effective training. 
In the medical domain, acquiring large-scale data is often challenging due to privacy concerns and the rarity of certain diseases. 
To address this data scarcity, we investigate the efficacy of training DNN models using synthetic data, generated based on domain-specific medical knowledge. 
Specifically, we develop a knowledge-driven Gaussian-composition synthesis algorithm for single-lead II ECGs, in which each heartbeat is represented by Gaussian-shaped P, Q, R, S, and T wave components. Using this simulator, we generate synthetic data for four abnormal electrocardiogram (ECG) classes: atrial fibrillation (AF), atrial flutter (AFLT), premature ventricular complex (PVC), and Wolff-Parkinson-White Syndrome (WPW). 
We evaluate the utility of this synthetic data by conducting abnormal ECG classification using ten different DNN architectures.
Our results demonstrate that synthetic-to-real training improves classification performance for three of the four target abnormalities, with the largest architecture-averaged gain of $33.2\%$ observed for AFLT.
Further analysis reveals that the performance enhancement from synthetic data is more pronounced with smaller real-world datasets. 
These findings suggest that domain-knowledge-based synthetic ECGs can serve as a useful pre-training resource, particularly in scenarios where real-world data are limited or difficult to obtain.

\keywords{Electrocardiogram  \and Data synthesis \and Deep learning.}
\end{abstract}
\section{Introduction}

Training Deep Neural Network (DNN) models demands a substantial volume of data, a resource that is not always readily accessible within the medical domain. 
When developing a disease classification model, having data related to the target disease is imperative. 
However, in instances involving rare diseases, gathering extensive datasets is challenging due to their infrequent occurrence. 
Furthermore, the integration and analysis of data collected from multiple healthcare facilities can be costly and complicated, often hindered by privacy concerns.
Given these constraints, there is a growing need for methods that enable the training of models under conditions where it is not feasible to amass extensive real-world patient data.

In tasks like object detection, DNN models trained with synthetic data have demonstrated their utility.
In object detection, a dataset was generated by synthesizing a 3D model of the target object and rendering it against random backgrounds, which was then employed for training DNN models \cite{tremblay2018training}.
In the context of human body part segmentation and depth estimation, the efficacy of synthesized datasets was demonstrated by \cite{varol17_surreal}.
In the data synthesis carried out in these studies, understanding the data synthesis targets, such as computer-generated representations of the target, plays a crucial role.

In the medical domain, decades of clinical and fundamental research have established a rich domain knowledge base, which can be leveraged to develop function-based synthesis methods. 
For instance, prior studies have resulted in sophisticated models capable of synthetically generating realistic electrocardiogram (ECG) signals~\cite{mcsharry2003dynamical,sayadi2010synthetic}. 
Unlike deep generative models, simulator-based synthesis can be conducted without training a generative model on patient-derived ECG recordings, as it relies on predefined functions and manually specified parameters rather than learning the signal distribution solely from data. 
However, existing ECG simulation studies have mainly focused on physiological signal generation, signal-processing tasks, or generic ECG modeling. 
The construction of labeled abnormal synthetic ECGs and their application to training DNN-based abnormal ECG classifiers under severe data scarcity remain underexplored.

This study tackles the challenge of classifying abnormal ECG classes with very limited real-world data by synthesizing class-specific abnormal ECGs. 
We first describe a simple Gaussian-composition ECG simulator that generates single-lead II ECG-like signals by superimposing P, Q, R, S, and T wave components. 
We then extend this base simulator with class-specific rules to synthesize four abnormal ECG classes: atrial fibrillation (AF), atrial flutter (AFLT), premature ventricular complex (PVC), and Wolff-Parkinson-White Syndrome (WPW). 
Subsequently, we train DNN models to classify each abnormal ECG class against normal ECG using synthesized data, real-world data, or synthetic-to-real training. Our work demonstrates the effectiveness and limitations of using class-specific synthesized ECG data for training classification models under severe data scarcity.

\section{Related Work}


Recent advancements in computer vision have explored synthetic data as a potent alternative to traditional natural image datasets. 
The work by \cite{KataokaIJCV2022,KataokaACCV2020} presents Formula-driven Supervised Learning, using the Fractal DataBase (FractalDB) of fractal-generated images to pretrain convolutional neural networks, enabling scalable dataset creation that sometimes surpasses traditional datasets in capturing unique visual features.
In \cite{tremblay2018training}, domain randomization is used to create synthetic images for training deep networks by randomizing parameters like lighting and pose, compelling the network to learn key object features and achieving competitive performance with minimal dependence on real-world data.
Additionally, \cite{varol17_surreal} presents a large-scale synthetic human dataset using the Skinned Multi-Person Linear Model (SMPL) body model, providing realistic renderings for various tasks such as pose estimation and optical flow. 
These studies collectively highlight the potential of synthetic data to enhance DNN training, reducing the dependency on large, annotated real-world datasets.

In the medical domain, efforts have been made to generate and synthesize data.
Several studies have used deep generative models, such as Generative Adversarial Networks (GANs), to produce chest X-ray images, addressing issues like class imbalance and the scarcity of labeled data \cite{salehinejad2018generalization,madani2018semi,koga2018general}.
In dermatology, GANs have been utilized to create skin images, demonstrating that these synthetic images can improve classification performance, especially for identifying rare and malignant conditions \cite{ghorbani2020dermgan}.
Another study developed a tumor segmentation model using a dataset of real CT images enhanced with synthetic tumors, achieving results on par with those trained on genuine images \cite{hu2023label}.
Beyond imaging, various techniques have been employed to generate synthetic electronic health records \cite{biswal2021eva,chintagunta2021medically,naseer2023scoehr}.
These methods are especially valuable given the high costs associated with acquiring real medical data.

In the field of ECG analysis, researchers have investigated a range of approaches to developing DNN models.
Studies have shown that DNNs can classify arrhythmias with accuracy comparable to physicians 
 \cite{hannun2019cardiologist}.
To address the scarcity of labeled data, ECG-specific augmentation techniques have been introduced to facilitate effective learning from limited datasets \cite{zhu2022geoecg,nonaka2021randecg,raghu2022data}.
Additionally, to address privacy concerns, techniques have been developed to transform collected data to retain disease characteristics while preventing the prediction of personal information like gender and age \cite{nolin2023privecg}.
Despite their proven effectiveness, DNNs in ECG analysis still face challenges, such as large-scale data collection and privacy concerns.


Various approaches have been developed to generate or synthesize ECG data.
For example, \cite{mcsharry2003dynamical} generated realistic ECG signals by coupling three ordinary differential equations, \cite{sayadi2010synthetic} produced arrhythmias by generating individual characteristic waves, and \cite{behar2014ecg} developed a simulator for maternal-foetal ECG activity mixtures.
In deep generative models, incorporating GANs with ordinary differential equations has proven effective for heart beat classification, as shown by \cite{golany2020simgans} and \cite{golany2021ecg}.
Furthermore, \cite{kaisti2023domain} used synthesized ECG data with varied waveform shapes, RR intervals, and noise levels to train a DNN for R-wave detection, while \cite{landajuela2022intracardiac} trained a DNN model on synthesized data to predict intracardiac electrical images from ECG signals.
In this study, we take a knowledge-driven approach and provide a self-contained description of a simple Gaussian-composition ECG simulator for generating synthetic normal and abnormal single-lead II ECGs. 
We use the synthesized ECGs as training data and evaluate whether class-specific synthetic signals can improve DNN-based abnormal ECG classification under data scarcity.

\section{Real-world ECG Data}

This study contrasts the classification performance of DNN models trained using synthesized data with those trained solely on real-world data, aiming to validate the efficacy of synthesized data in ECG classification.
For the real-world data, we prepared dataset by combining samples from PTB-XL dataset \cite{wagner2020ptb,goldberger2000physiobank} employed as normal or abnormal state ECG.
The PTB-XL dataset consists of $21,799$ samples of 12-lead ECGs from $18,869$ subjects collected in Germany between October 1989 and June 1996 using devices from Schiller AG.
Each sample has a duration of $10$ seconds, with a sampling frequency of $500$ Hz, and is associated with at least one of a total of 71 distinct statements, providing information relevant to disease classification.
In this study, we utilized lead-II ECG of $7,185$ samples labeled as `NORM', $1,514$ samples labeled as `AFIB', $56$ samples labeled as `AFLT', $1,030$ samples labeled as `PVC', and $71$ samples labeled as `WPW'.

\section{Method: Knowledge-driven Gaussian-composition synthesis of abnormal ECGs}

The ECG possesses fundamental characteristics that are widely used in clinical interpretation and signal-processing research. A typical heartbeat consists of prominent waveform components, namely the P wave, QRS complex, and T wave, which can be used as a basis for knowledge-driven data synthesis. 
In this study, we synthesize single-lead II ECG signals using a simple Gaussian-composition simulator. The simulator is designed to generate diverse ECG-like waveforms for model training and data augmentation, rather than to serve as a clinically complete physiological simulator.

Each heartbeat is represented as the sum of five Gaussian-shaped wave components corresponding to the P, Q, R, S, and T waves. Let $w \in \{P,Q,R,S,T\}$ denote a waveform component. Each component is parameterized by a signed amplitude $a_w$, temporal shift $\mu_w$, and width $\sigma_w$. The waveform of one beat is computed as
\[
x_{\mathrm{beat}}(t)
=
\sum_{w \in \{P,Q,R,S,T\}}
a_w
\exp \left(
-\frac{1}{2}
\left(
\frac{t-\mu_w}{\sigma_w}
\right)^2
\right).
\]
The temporal shift and width parameters are defined in normalized beat time. By changing the amplitude, shift, and width of each component, the simulator can generate diverse P-QRS-T morphologies.

To introduce diversity, we use a two-level perturbation scheme. 
For each synthetic recording, the initial P, Q, R, S, and T wave parameters are sampled from Gaussian distributions centered at manually specified base values. 
This sample-level perturbation introduces inter-sample variability. After each beat is generated, smaller Gaussian perturbations are added to the parameters before generating the next beat, introducing beat-to-beat variability within the same recording. 
The beat-level perturbations are intentionally smaller than the sample-level perturbations so that each recording remains internally consistent while the dataset covers diverse ECG morphologies. The detailed parameterization is provided in Appendix~\ref{app:ecg_synthesis}.

Beats are concatenated until the signal length exceeds 10 seconds at $500$ Hz, and the signal is then trimmed to $5,000$ time steps to match the PTB-XL recordings used in this study. 
We further add white noise and sinusoidal baseline fluctuation to simulate measurement noise and baseline wander. 
Before being provided to neural networks, each ECG sample is standardized by subtracting its mean and dividing by its standard deviation.

Normal synthetic ECGs are generated using this base simulator. Abnormal synthetic ECGs are generated by applying class-specific modifications to the same waveform representation. Specifically, our investigation focuses on four prevalent abnormal ECG patterns: atrial fibrillation (AF), atrial flutter (AFLT), premature ventricular complex (PVC), and Wolff-Parkinson-White Syndrome (WPW). These conditions exhibit distinct characteristics within the ECG waveform. While AF and PVC are frequently encountered in clinical settings, instances of AFLT and WPW are comparatively rare. Hereafter, we explain the specific procedures used to synthesize each abnormal ECG class.

\begin{figure}[htbp]
  \centering
  \subfigure[Real-world]{%
    \includegraphics[width=.475\linewidth]{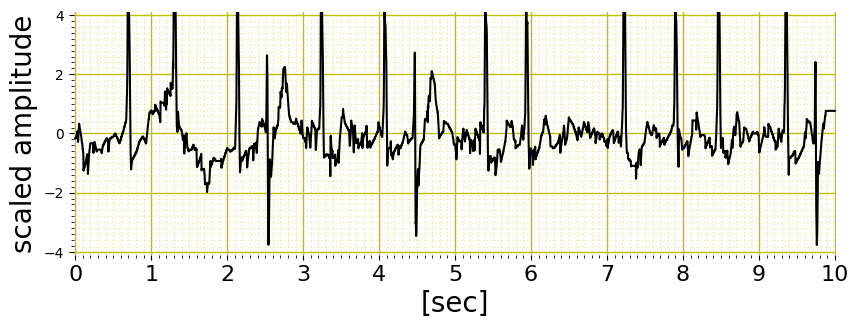}%
    \label{fig:real-af}
  }
  \hfill
  \subfigure[Synthesized]{%
    \includegraphics[width=.475\linewidth]{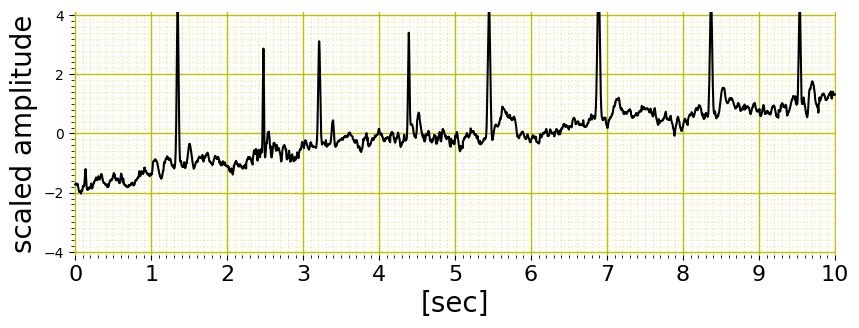}%
    \label{fig:syn-af}
  }
  \caption{Examples of real-world and synthesized ECG of AF.}
  \label{fig:af}
\end{figure}

For the synthesis of AF ECGs, we used the following rule.
AF is characterized by the loss of regular atrial excitation, resulting in fine trembling of the atrial muscles. 
This condition manifests in the waveform as the absence of the P wave, oscillations in the baseline, and a reduction in the RR interval. 
Consequently, in the AF synthesis algorithm, the P wave is omitted, and a repetitive waveform mimicking the baseline oscillations is generated, while the cycle duration is shortened. The synthesis of the QRST components follows the same approach as in a normal ECG.
Resulting synthesized AF ECG sample is visualized in Figure~\ref{fig:af} along with real-world AF data.

\begin{figure}[htbp]
  \centering
  \subfigure[Real-world]{%
    \includegraphics[width=.475\linewidth]{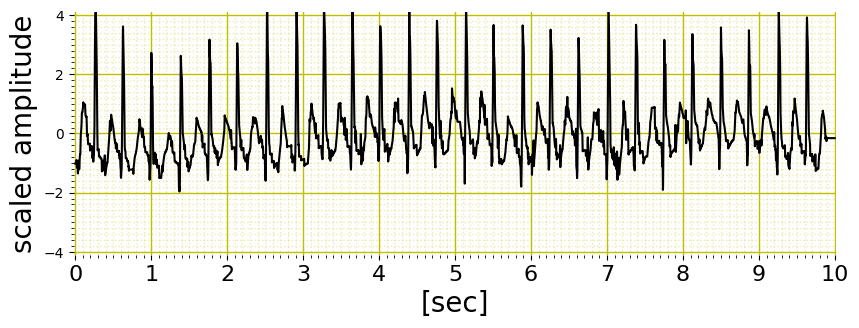}%
    \label{fig:real-aflt}
  }
  \hfill
  \subfigure[Synthesized]{%
    \includegraphics[width=.475\linewidth]{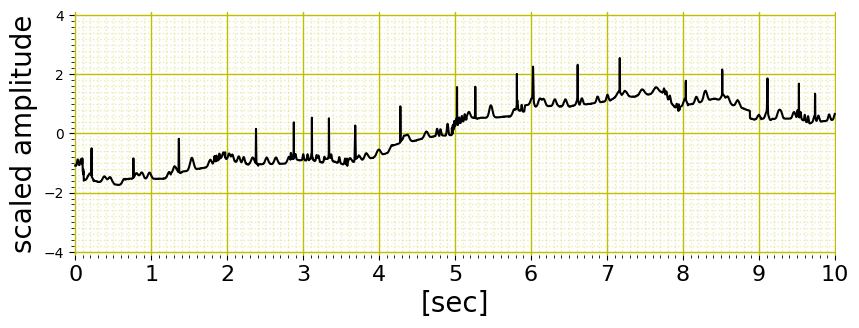}%
    \label{fig:syn-aflt}
  }
  \caption{Examples of real-world and synthesized ECG of AFLT.}
  \label{fig:aflt}
\end{figure}

For the synthesis of AFLT ECGs, we used the following rule.
AFLT is a condition where the atria depolarize rapidly and regularly at approximately 300 beats per minute, leading to abnormal heart rhythms. 
The characteristic waveform of AFLT lacks the P wave and displays a sawtooth-like flutter wave. 
To replicate these features in the synthesis algorithm, the P wave is omitted, and a repetitive waveform mimicking the baseline oscillations similar to those seen in AF is generated. 
The synthesis of the QRS complex is conducted in the same manner as for a normal ECG.
Resulting synthesized AFLT ECG sample is visualized in Figure~\ref{fig:aflt} along with corresponding real-world data.

\begin{figure}[htbp]
  \centering
  \subfigure[Real-world]{%
    \includegraphics[width=.475\linewidth]{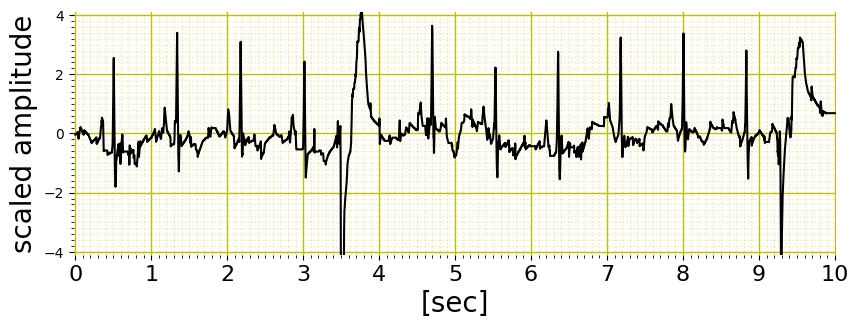}%
    \label{fig:real-pvc}
  }
  \hfill
  \subfigure[Synthesized]{%
    \includegraphics[width=.475\linewidth]{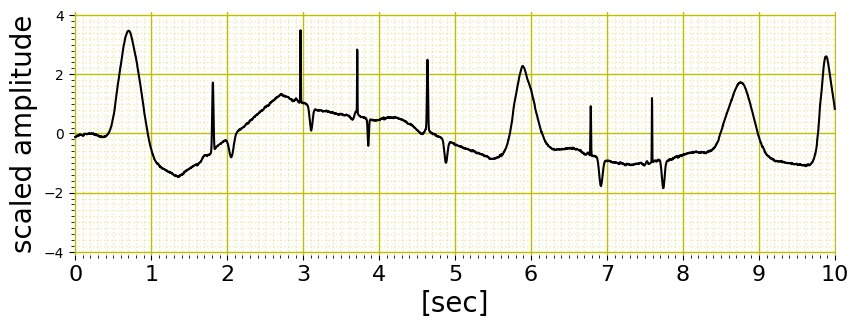}%
    \label{fig:syn-pvc}
  }
  \caption{Examples of real-world and synthesized ECG of PVC.}
  \label{fig:pvc}
\end{figure}

For the synthesis of PVC ECGs, we used the following rule.
PVC occur when an ectopic excitation stimulates the heart ahead of the normal contraction. 
This condition is characterized by the absence of a preceding P wave and an increased width of the QRS complex compared to normal. 
To reflect these characteristics, the synthesis algorithm generates PVC heartbeats with a diminished P wave and a broadened QRS complex according to predefined parameters. 
All other heartbeats are synthesized in the same manner as normal ECG.
Resulting synthesized PVC ECG sample is visualized in Figure~\ref{fig:pvc} along with corresponding real-world data.

\begin{figure}[htbp]
  \centering
  \subfigure[Real-world]{%
    \includegraphics[width=.475\linewidth]{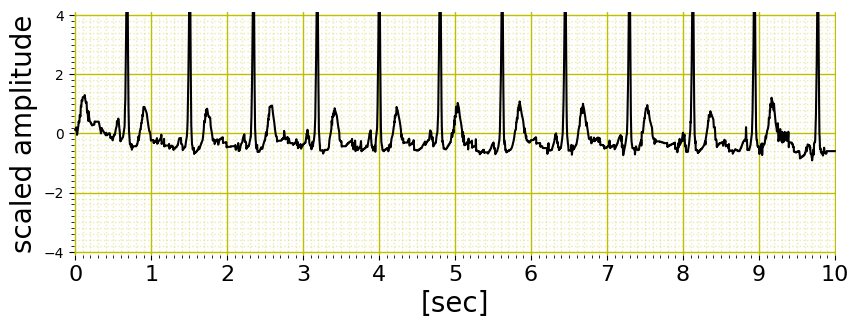}%
    \label{fig:real-wpw}
  }
  \hfill
  \subfigure[Synthesized]{%
    \includegraphics[width=.475\linewidth]{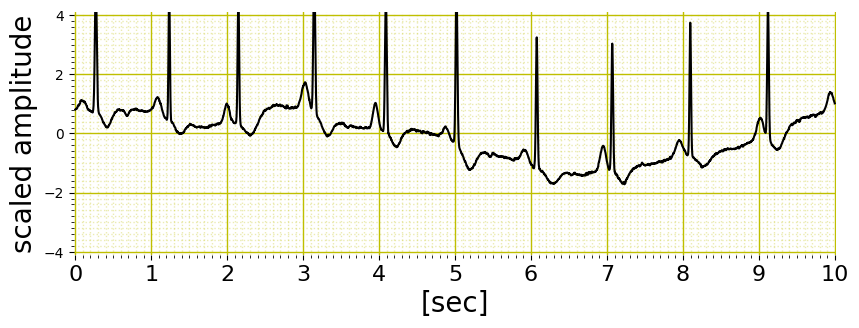}%
    \label{fig:syn-wpw}
  }
  \caption{Examples of real-world and synthesized ECG of WPW.}
  \label{fig:wpw}
\end{figure}

For the synthesis of WPW ECGs, we used the following rule.
WPW is caused by the premature ventricular excitation via an accessory pathway known as the Kent bundle, which connects the atria and ventricles and precedes normal ventricular excitation. 
The characteristic features observed in the ECG include a shortened PQ interval, the presence of a delta wave that replaces the QRS complex with a gradual upslope, and an extended QRS duration. 
To replicate these features, the synthesis algorithm has been extended to generate a delta wave in addition to the PQRST waves present in a normal ECG.
Resulting synthesized WPW ECG sample is visualized in Figure~\ref{fig:wpw} along with corresponding real-world data.


\section{Experiment}

To evaluate the effectiveness of training with synthesized ECG data, we conducted a comparative analysis of abnormal ECG classification models across four types of abnormalities, comparing three training scenarios: (1) using only real-world data (``Real''), (2) using only synthesized data (``Syn''), and (3) fine-tuning a model pre-trained on synthesized data with real-world data (``Syn$\rightarrow$Real'').

\subsection{Model architectures and data split}

We conducted our experiments with various categories of DNN architectures, including CNN-based architectures, RNN-based architectures, and Transformer and its variants.
As for CNN-based architectures, we examined ResNet-18, ResNet-34, ResNet50 \cite{he2016deep} and EfficientNet (B0) \cite{tan2019efficientnet}.\footnote{
We convert 2D convolutional and pooling layers to its 1D counterparts following \cite{nonaka2021depth}.
}
In addition to the CNN-based architectures, we incorporated RNN-based architectures, including LSTM \cite{hochreiter1997long} and GRU \cite{chung2014empirical}.
Furthermore, we evaluated Transformer \cite{vaswani2017attention} along with LUNA transformer \cite{ma2021luna}, S4 \cite{gu2021efficiently}, and MEGA \cite{ma2022mega}, known for their effective performance in handling long sequence tasks.
In summary, we selected ten DNN architectures and trained each architecture for classification of four abnormal ECGs.

\begin{table}[htbp]
    \caption{Sample size of each dataset.}
    \label{tab:n_data_ds}
    \centering
    \begin{tabularx}{\textwidth}{X R R R R R}
    \toprule
    \multirow{2}{*}{Class} 
        & \multicolumn{3}{c}{PTB-XL (Real-world)} 
        & \multicolumn{2}{c}{Synthesized} \\
    
        & Train & Val. & Test 
        & Train & Val. \\
    \midrule
        AF &$968$ & $243$ & $303$ & $5,000$ & $2,500$ \\
        AFLT &$35$ & $9$ & $12$ & $5,000$ & $2,500$ \\
        PVC &$659$ & $165$ & $206$ & $5,000$ & $2,500$ \\
        WPW &$44$ & $12$ & $15$ & $5,000$ & $2,500$ \\
        Normal & $4,598$ & $1,150$ & $1,437$ & $5,000$ & $2,500$ \\
    \bottomrule
    \end{tabularx}
\end{table}

The datasets used in our experiments were split according to following procedure.
For the real-world dataset, PTB-XL dataset, the dataset was first randomly split into train/val and test sets at a ratio of $8:2$.
The train/val set was further divided into six independent training and validation sets, each with an 8:2 ratio.
For the synthesized data, we generated six independent training and validation sets. 
For each synthetic class and each split, we generated $5,000$ training samples and $2,500$ validation samples, as shown in Table~\ref{tab:n_data_ds}. 
For each binary classification task, we constructed balanced synthetic datasets by combining abnormal samples as the positive class and normal samples as the negative  class, resulting in $10,000$ training samples and $5,000$ validation samples per split.

\subsection{Model training settings}

Regarding the classification of abnormal state ECGs, we adopted the following approach. 
We maintained the partition of the training, validation, and test sets both before and after combining the datasets. 
Samples of abnormal state ECGs from the PTB-XL or synthesized abnormal ECG datasets were labeled as positive instances, while samples of normal ECGs from the PTB-XL and synthesized normal ECG datasets were labeled as negative instances. 

\subsubsection{Training model with full data}

For the models trained under ``Real'' setting, we took the following steps: To mitigate the impact of imbalance between positive and negative samples, the inverse of the ratio of positive to negative samples in the training set was employed as weights for the positive samples. 
A single pair of training and validation sets was employed to search for the optimal hyperparameters. 
After selecting the optimal hyperparameters, five independent training runs were carried out using the remaining five training and validation set pairs.
Ultimately, each model from the five independent training runs was applied to a common test set, and the average of their $F1$-scores was used as the final evaluation metric.

For the models trained under ``Syn'' setting, the following steps were taken: A total of $10,000$ samples were used for training, with $5,000$ for positive class samples and $5,000$ for negative class samples. 
Similar to the ``Real'' setting, a single pair of train and validation set was used to search for hyperparameters, and five independent training runs were conducted with five train and validation set pairs under the specified hyperparameters. 
Following this, the performance of all five models was evaluated using the test set from the corresponding real-world dataset and mean $F1$-score was calculated, respectively.

For the models of ``Syn $\rightarrow$ Real'' settings, we fine-tuned the models trained under ``Syn'' setting with the corresponding real-world data.
Consistent with the models of ``Real'' setting, we used the inverse of the ratio of positive to negative samples in the training set as weights for the positive samples, and optimal hyperparameters were searched using a single pair of training and validation set.
After selecting the optimal hyperparameters, five independent training runs were carried out using the remaining five training and validation set pairs and evaluated the performance using test set $F1$-score as a metric.

As for preprocessing, we applied scaling by subtracting the mean value from each sample and dividing by the standard deviation. 
As for data augmentation, we applied random shifting and random masking with rates of shift and masking determined through hyper-parameter search. 
The batch size was set to $512$, and the maximum number of epochs was set to $500$, with validation conducted every five epochs.
Early stopping was applied if the validation loss did not improve for five consecutive evaluations. 
We used Adam \cite{kingma2014adam} as the optimizer with the learning rate determined by hyper-parameter search.
Regarding the loss function, we used binary cross entropy loss.
We computed F1-scores using a fixed decision threshold of $0.5$.

\subsubsection{Training model with deliberately reduced positive class training set}

To assess how synthesized data impacts model performance when real-world data is limited, we trained models using varying numbers of abnormal ECG samples from the real-world dataset.
All negative samples in the PTB-XL dataset were used for training, while the number of positive samples was constrained to $1,000$, $500$, $250$, $100$, $50$, $25$, $10$, $5$, $2$, and $1$.\footnote{
If the number of samples in the training set was smaller than the constraints, we used training samples.
}
In this experiment, hyperparameters remained consistent with the previous setting.
Five independent trials were conducted, and each model obtained was applied to a common test set. 
The average of their $F1$-scores was then used as the final evaluation.

\section{Result}

In this section, we present the obtained experimental results for abnormal ECG classification experiments.
We first present result for abnormal ECG classification with three different training data setting, namely ``Real'', ``Syn'' and ``Syn $\rightarrow$ Real'', for four abnormal ECG classes respectively.
Subsequently, we present analysis on performance of ``Syn $\rightarrow$ Real'' setting over ``Real'' setting under deliberately downsized training dataset.
Furthermore, to analyze the relationship between the amount of real-world data used during training and the performance improvement observed with the ``Syn $\rightarrow$ Real'' approach, we conducted experiments by deliberately reducing the training data for the positive class.

\subsection{Classification performance comparison}

\begin{table}[h!]
\centering
\caption{
AFLT classification performance ($F1$-score) 
}\label{tab01:aflt}
    \begin{threeparttable}
    \begin{tabularx}{\textwidth}{
        X C C C D
    }
    \toprule
     & $a.$ Real & $b.$ Syn & $c.$ Syn $\rightarrow$ Real & Gain (\%)\tnote{*}\\
    \midrule
    $n$ synthesized\tnote{\dag} 
        & \multicolumn{1}{r}{$0\quad\quad$} 
        & \multicolumn{1}{r}{$10,000\quad\quad$} 
        & \multicolumn{1}{r}{$10,000\quad\quad$}  & - \\
    $n$ real-world\tnote{\dag} 
        & \multicolumn{1}{r}{$4,633\quad\quad$} 
        & \multicolumn{1}{r}{$0\quad\quad$} 
        & \multicolumn{1}{r}{$4,633\quad\quad$}  & - \\
    \midrule
    EfficientNet-B0
        &$0.9032\pm0.0784$
        &$0.0745\pm0.0461$
        &$0.9840\pm0.0196$ 
        & \multicolumn{1}{r}{$8.946\quad$} \\
    GRU
	&$0.6954\pm0.2071$
	&$0.1248\pm0.0434$
	&$0.8407\pm0.0778$
	& \multicolumn{1}{r}{$20.894\quad$} \\
    LSTM
        &$0.8456\pm0.0719$
	&$0.0796\pm0.0327$
	&$0.8258\pm0.0891$
	& \multicolumn{1}{r}{$-2.342\quad$} \\
    Luna
	&$0.6101\pm0.2384$
	&$0.0802\pm0.0267$
	&$0.7894\pm0.1204$
	& \multicolumn{1}{r}{$29.389\quad$} \\
    Mega
	&$0.6737\pm0.2143$
	&$0.0602\pm0.0478$
	&$0.8166\pm0.1260$
	& \multicolumn{1}{r}{$21.211\quad$} \\
    ResNet18
	&$0.7145\pm0.3034$
	&$0.0575\pm0.0451$
	&$0.8909\pm0.1623$
	& \multicolumn{1}{r}{$24.689\quad$} \\
    ResNet34
        &$0.5884\pm0.3334$
        &$0.0947\pm0.1034$
        &$0.9252\pm0.1111$
        & \multicolumn{1}{r}{$57.240\quad$} \\
    ResNet50
        &$0.3961\pm0.2252$
        &$0.0404\pm0.0112$
        &$0.8280\pm0.1122$
        & \multicolumn{1}{r}{$109.038\quad$} \\
    S4	
        &$0.5973\pm0.2701$
        &$0.0867\pm0.0661$
        &$0.8106\pm0.1600$
        & \multicolumn{1}{r}{$35.711\quad$} \\
    Transformer
	&$0.6250\pm0.2456$
	&$0.0622\pm0.0477$
	&$0.7964\pm0.1164$
	& \multicolumn{1}{r}{$27.424\quad$} \\
    \midrule
    Average
        &$0.6649$
        &$0.0761$
        &$0.8508$
        & \multicolumn{1}{r}{$33.220\quad$} \\ 
    \bottomrule
    \end{tabularx}
  \begin{tablenotes}
    \small
    \item[*] The relative improvement from ``Real'' to ``Syn$\rightarrow$Real'', calculated as $(c - a) / a \times 100$.
    \item[\dag] The total number of synthesized and real-world data used for training the model, respectively.
    \end{tablenotes}
\end{threeparttable}
\end{table}

The results of classifying AFLT state ECG from normal ECG are presented in Table~\ref{tab01:aflt}.
For the model trained under ``Real'' setting, the EfficientNet-B0 achieved the best performance, with a mean $F1$-score of $0.9032$.
In contrast, when trained under ``Syn'' setting, the GRU outperformed others, albeit with a significantly lower mean $F1$-score of $0.1248$.
Interestingly, with ``Syn $\rightarrow$ Real'' setting, the EfficientNet-B0 architecture achieved the best performance, with an $F1$-score of $0.9840$.
On average, across the ten models, the ``Syn $\rightarrow$ Real'' approach demonstrated a $33.22\%$ improvement over the models trained exclusively with real-world data (Classification result evaluated with AUPRC are shown in Appendix~\ref{app:auprc}, Table~\ref{tab01a:aflt}).

\begin{table}[h!]
\centering
\caption{
WPW classification performance ($F1$-score)
}\label{tab01:wpw}
    \begin{threeparttable}
    \begin{tabularx}{\textwidth}{
        X C C C D
    }
    \toprule
     & $a.$ Real & $b.$ Syn & $c.$ Syn $\rightarrow$ Real & Gain (\%)\tnote{*}\\
    \midrule
    $n$ synthesized\tnote{\dag} 
        & \multicolumn{1}{r}{$0\quad\quad$} 
        & \multicolumn{1}{r}{$10,000\quad\quad$} 
        & \multicolumn{1}{r}{$10,000\quad\quad$}  & - \\
    $n$ real-world\tnote{\dag} 
        & \multicolumn{1}{r}{$4,642\quad\quad$} 
        & \multicolumn{1}{r}{$0\quad\quad$} 
        & \multicolumn{1}{r}{$4,642\quad\quad$}  & - \\
    \midrule
    EfficientNet-B0
        &$0.1242\pm0.1078$
        &$0.0235\pm0.0140$
        &$0.2021\pm0.1237$
        & \multicolumn{1}{r}{$62.721\quad$} \\
    GRU
	&$0.1159\pm0.0295$
	&$0.0484\pm0.0079$
	&$0.1658\pm0.0327$
	& \multicolumn{1}{r}{$43.054\quad$} \\
    LSTM
	&$0.1841\pm0.0353$
	&$0.0466\pm0.0173$
	&$0.2134\pm0.0937$
	& \multicolumn{1}{r}{$15.915\quad$} \\
    Luna
	&$0.0797\pm0.0387$
	&$0.0213\pm0.0033$
	&$0.0636\pm0.0530$
        & \multicolumn{1}{r}{$-20.201\quad$} \\
    Mega
	&$0.2223\pm0.1432$
	&$0.0494\pm0.0135$
	&$0.2161\pm0.0844$
	& \multicolumn{1}{r}{$-2.789\quad$} \\
    ResNet18
	&$0.2181\pm0.1130$
	&$0.0321\pm0.0104$
	&$0.2958\pm0.0777$
	& \multicolumn{1}{r}{$35.626\quad$} \\
    ResNet34
        &$0.1562\pm0.0439$
        &$0.0565\pm0.0466$
        &$0.2253\pm0.0352$
        & \multicolumn{1}{r}{$44.238\quad$} \\
    ResNet50
        &$0.0817\pm0.0783$
        &$0.0154\pm0.0096$
        &$0.1327\pm0.0465$
        & \multicolumn{1}{r}{$62.424\quad$} \\
    S4	
        &$0.1389\pm0.0538$
        &$0.0558\pm0.0091$
        &$0.1780\pm0.0233$
        & \multicolumn{1}{r}{$28.150\quad$} \\
    Transformer
	&$0.1405\pm0.0632$
	&$0.0819\pm0.0869$
	&$0.1961\pm0.0621$
	& \multicolumn{1}{r}{$39.573\quad$} \\
    \midrule
    Average
        & $0.1462$
        & $0.0431$
        & $0.1889$
        & \multicolumn{1}{r}{$30.871\quad$} \\    
    \bottomrule
    \end{tabularx}
    \begin{tablenotes}
    \small
    \item[*] The relative improvement from ``Real'' to ``Syn$\rightarrow$Real'', calculated as $(c - a) / a \times 100$.
    \item[\dag] The total number of synthesized and real-world data used for training the model, respectively.
    \end{tablenotes}
\end{threeparttable}
\end{table}

The results of classifying WPW state ECG from normal ECG are presented in Table~\ref{tab01:wpw}.
For the model trained under ``Real'' setting, the Mega achieved the best performance, with a mean $F1$-score of $0.2223$.
In contrast, when trained under ``Syn'' setting, the Transformer outperformed others, albeit with a significantly lower mean $F1$-score of $0.0819$.
Interestingly, with ``Syn $\rightarrow$ Real'' setting, the ResNet-18 architecture achieved the best performance, with an $F1$-score of $0.2958$.
On average, across the ten models, the ``Syn $\rightarrow$ Real'' approach demonstrated a $30.87\%$ improvement over the models trained exclusively with real-world data (Classification result evaluated with AUPRC are shown in Appendix~\ref{app:auprc}, Table~\ref{tab01a:wpw}).

\begin{table}[h!]
\centering
\caption{
PVC classification performance ($F1$-score) 
}\label{tab01:pvc}
    \begin{threeparttable}
    \begin{tabularx}{\textwidth}{
        X C C C D
    }
    \toprule
     & $a.$ Real & $b.$ Syn & $c.$ Syn $\rightarrow$ Real & Gain (\%)\tnote{*}\\
    \midrule
    $n$ synthesized\tnote{\dag} 
        & \multicolumn{1}{r}{$0\quad\quad$} 
        & \multicolumn{1}{r}{$10,000\quad\quad$} 
        & \multicolumn{1}{r}{$10,000\quad\quad$}  & - \\
    $n$ real-world\tnote{\dag} 
        & \multicolumn{1}{r}{$5,257\quad\quad$} 
        & \multicolumn{1}{r}{$0\quad\quad$} 
        & \multicolumn{1}{r}{$5,257\quad\quad$}  & - \\
    \midrule
    EfficientNet-B0	
        &$0.8698\pm0.0327$
        &$0.2368\pm0.1292$
        &$0.9278\pm0.0077$
        & \multicolumn{1}{r}{$6.668\quad$} \\    
    GRU
	&$0.9031\pm0.0223$
	&$0.2150\pm0.0417$
	&$0.9281\pm0.0170$
	& \multicolumn{1}{r}{$2.768\quad$} \\    
    LSTM
	&$0.9024\pm0.0198$
	&$0.2576\pm0.0484$
	&$0.9292\pm0.0050$
	& \multicolumn{1}{r}{$2.970\quad$} \\    
    Luna
	&$0.8022\pm0.0386$
	&$0.3084\pm0.0598$
	&$0.8170\pm0.0171$
	& \multicolumn{1}{r}{$1.845\quad$} \\    
    Mega
	&$0.8957\pm0.0153$
	&$0.3616\pm0.0236$
	&$0.9038\pm0.0164$
	& \multicolumn{1}{r}{$0.904\quad$} \\    
    ResNet18
	&$0.8485\pm0.0475$
	&$0.1240\pm0.1105$
	&$0.9058\pm0.0156$
	& \multicolumn{1}{r}{$6.753\quad$} \\    
    ResNet34
        &$0.8679\pm0.0203$
        &$0.0993\pm0.0455$
        &$0.9127\pm0.0210$
        & \multicolumn{1}{r}{$5.162\quad$} \\    
    ResNet50	
        &$0.8068\pm0.0728$
        &$0.0885\pm0.0599$
        &$0.8706\pm0.0088$
        & \multicolumn{1}{r}{$7.908\quad$} \\    
    S4
        &$0.8818\pm0.0163$	
        &$0.2649\pm0.0188$	
        &$0.9011\pm0.0167$
        & \multicolumn{1}{r}{$2.189\quad$} \\  
    Transformer
	&$0.7860\pm0.0279$
	&$0.2655\pm0.0457$
	&$0.8671\pm0.0127$
	& \multicolumn{1}{r}{$10.318\quad$} \\    
    \midrule
    Average
        & $0.8564$	
        &$0.2222$	
        &$0.8963$	
        & \multicolumn{1}{r}{$4.749\quad$} \\    
    \bottomrule
    \end{tabularx}
  \begin{tablenotes}
    \small
    \item[*] The relative improvement from ``Real'' to ``Syn$\rightarrow$Real'', calculated as $(c - a) / a \times 100$.
    \item[\dag] The total number of synthesized and real-world data used for training the model, respectively.
    \end{tablenotes}
\end{threeparttable}
\end{table}

The results of classifying PVC state ECG from normal ECG are presented in Table~\ref{tab01:pvc}.
For the model trained under ``Real'' setting, the GRU achieved the best performance, with a mean $F1$-score of $0.9031$.
In contrast, when trained under ``Syn'' setting, the Mega outperformed others, albeit with a significantly lower mean $F1$-score of $0.3616$.
Interestingly, with ``Syn $\rightarrow$ Real'' setting, the LSTM architecture achieved the best performance, with an $F1$-score of $0.9292$.
On average, across the ten models, the ``Syn $\rightarrow$ Real'' approach demonstrated a $4.75\%$ improvement over the models trained exclusively with real-world data (Classification result evaluated with AUPRC are shown in Appendix~\ref{app:auprc}, Table~\ref{tab01a:pvc}).

\begin{table}[h!]
\centering
\caption{
AF classification performance ($F1$-score)
}\label{tab01:af}
    \begin{threeparttable}
    \begin{tabularx}{\textwidth}{
        X C C C D
    }
    \toprule
     & $a.$ Real & $b.$ Syn & $c.$ Syn $\rightarrow$ Real & Gain (\%)\tnote{*}\\
    \midrule
    $n$ synthesized\tnote{\dag} 
        & \multicolumn{1}{r}{$0\quad\quad$} 
        & \multicolumn{1}{r}{$10,000\quad\quad$} 
        & \multicolumn{1}{r}{$10,000\quad\quad$}  & - \\
    $n$ real-world\tnote{\dag} 
        & \multicolumn{1}{r}{$5,566\quad\quad$} 
        & \multicolumn{1}{r}{$0\quad\quad$} 
        & \multicolumn{1}{r}{$5,566\quad\quad$}  & - \\
    \midrule
    EfficientNet-B0	
        &$0.9560\pm0.0068$
        &$0.4173\pm0.0312$
        &$0.9459\pm0.0053$
	& \multicolumn{1}{r}{$-1.068\quad$} \\    
    GRU
	&$0.9452\pm0.0140$
        &$0.7175\pm0.0121$
        &$0.9559 \pm 0.0128$
        & \multicolumn{1}{r}{$1.119\quad$} \\    
    LSTM
	&$0.9375\pm0.0446$
        &$0.4658\pm0.3601$
        &$0.9349\pm0.0322$
        & \multicolumn{1}{r}{$-0.278\quad$} \\    
    Luna
	&$0.9327\pm0.0081$
        &$0.5268\pm0.0134$
        &$0.9043\pm0.0131$
        & \multicolumn{1}{r}{$-3.141\quad$} \\    
    Mega
	&$0.9285\pm0.0085$
        &$0.6680\pm0.0339$
        &$0.8258\pm0.2647$
        & \multicolumn{1}{r}{$-12.436\quad$} \\    
    ResNet18
	&$0.9522\pm0.0064$
        &$0.2850\pm0.0790$
        &$0.9466\pm0.0097$
        & \multicolumn{1}{r}{$-0.592\quad$} \\    
    ResNet34
        &$0.9559\pm0.0084$
        &$0.2797\pm0.0751$
        &$0.9376\pm0.0085$
	& \multicolumn{1}{r}{$-1.952\quad$} \\    
    ResNet50
        &$0.9565\pm0.0068$
        &$0.3875\pm0.0790$
        &$0.9493\pm0.0077$
	& \multicolumn{1}{r}{$-0.758\quad$} \\    
    S4
        &$0.9314\pm0.0083$
        &$0.6021\pm0.0352$
        &$0.9596\pm0.0158$
	& \multicolumn{1}{r}{$2.939\quad$} \\    
    Transformer
	&$0.9311\pm0.0171$
        &$0.5880\pm0.0306$
        &$0.8952\pm0.0156$
        & \multicolumn{1}{r}{$-4.010\quad$} \\    
    \midrule
    Average 
        &$0.9427$
        &$0.4938$
        &$0.9255$
        & \multicolumn{1}{r}{$-2.018\quad$} \\    
    \bottomrule
    \end{tabularx}
  \begin{tablenotes}
    \small
    \item[*] The relative improvement from ``Real'' to ``Syn$\rightarrow$Real'', calculated as $(c - a) / a \times 100$.
    \item[\dag] The total number of synthesized and real-world data used for training the model, respectively.    
    \end{tablenotes}
\end{threeparttable}
\end{table}

The results of classifying AF state ECG from normal ECG are presented in Table~\ref{tab01:af}.
For the model trained under ``Real'' setting,  the EfficientNet-B0 achieved the best performance, with a mean $F1$-score of $0.9560$.
In contrast, when trained under ``Syn'' setting, the GRU outperformed others, albeit with a significantly lower mean $F1$-score of $0.7175$.
With ``Syn $\rightarrow$ Real'' setting, the S4 architecture achieved the best performance, with an $F1$-score of $0.9596$.
On average, across the ten models, the ``Syn $\rightarrow$ Real'' approach demonstrated a $2.02\%$ deterioration over the models trained exclusively with real-world data (Classification result evaluated with AUPRC are shown in Appendix~\ref{app:auprc}, Table~\ref{tab01a:af}).

\subsection{Analysis of performance under downsized real-world data}

\begin{figure}
    \centering
    \includegraphics[width=\textwidth]{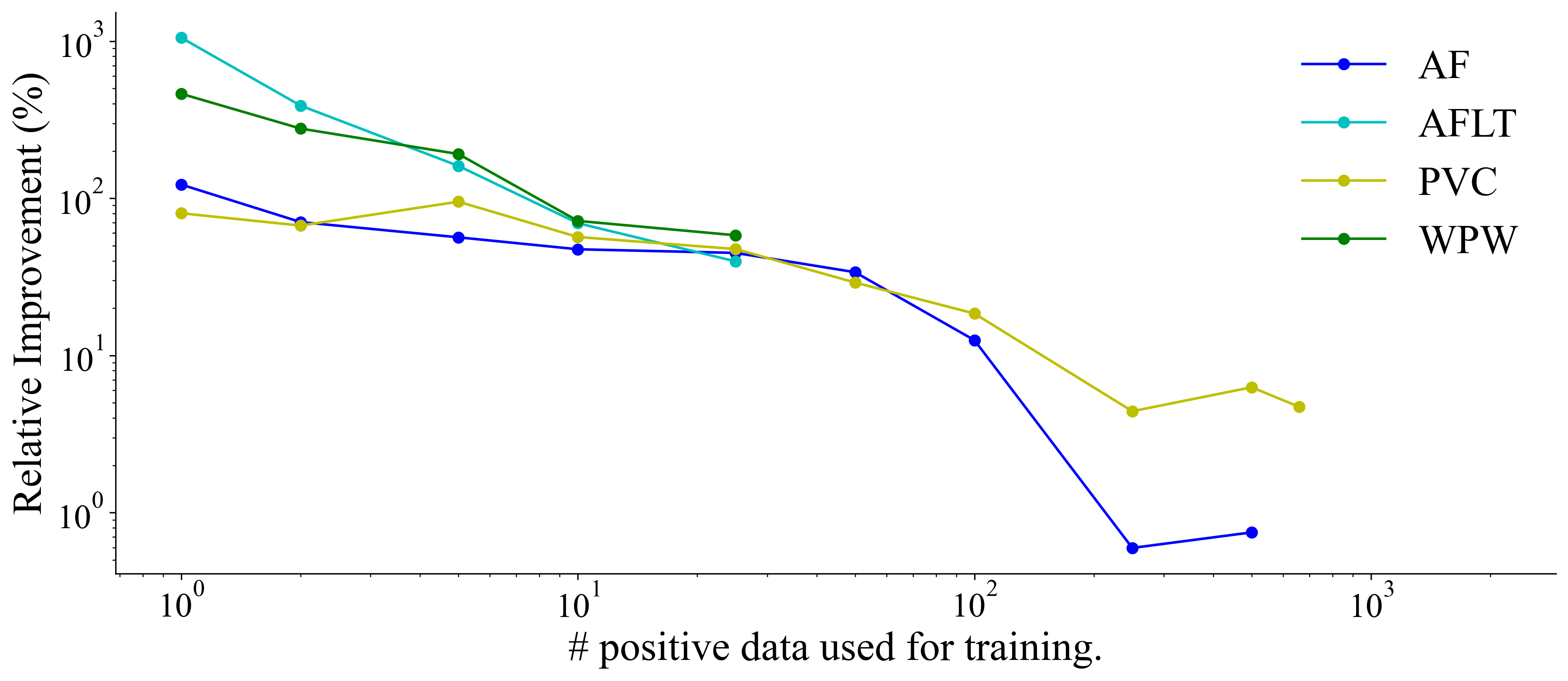}
    \caption{
Average improvement of the ``Syn $\rightarrow$ Real'' approach over the models trained with ``Real'' setting.
For all four abnormal ECG classification task, improvement rate increases with the decrease of the real-world data.
We did not plot point if the relative improvement was negative (= AF with all positive samples).
} \label{fig1}
\end{figure}

Subsequently, we examined the classification performance under deliberately reduced number of positive class training data.
We calculated relative performance improvement of ``Syn $\rightarrow$ Real'' setting over ``Real'' setting for each number of positive class training data for all four abnormal ECGs, respectively.
The result is shown in Figure~\ref{fig1}.
Although rate of improvement differs among four abnormal ECG classification task conducted, we observed similar trend of increasing improvement with a decrease of the real-world data.

\section{Discussion}

\begin{table}[h!]
\centering
\caption{
Number of positive data used during training and average improvement.
}\label{tab01:ndata}
    \begin{tabularx}{\textwidth}{E R R R R}
    \hline
         & AFLT & WPW & PVC & AF \\
    \hline
    \# positive data during training
        & $35$
        & $44$
        & $659$
        & $968$
    \\
    Average gain (\%)
        & $33.220$
        & $30.871$
        & $4.749$        
        & $-2.018$
    \\
    \hline
    \end{tabularx}
\end{table}

In the classification tasks of three abnormal ECG categories, excluding AF, an improvement in classification performance was observed under the ``Syn $\rightarrow$ Real'' setting. 
As shown in Table~\ref{tab01:ndata}, the performance improvement with ``Syn $\rightarrow$ Real'' over the ``Real'' setting was more pronounced with fewer positive class training samples. 
This trend is further corroborated by Figure~\ref{fig1}, where the average improvement rate under the ``Syn $\rightarrow$ Real'' setting increases as the number of positive class data points is deliberately reduced. 
This finding indicates that as the availability of real-world data decreases, the benefit of pre-training with synthetic data becomes more substantial.
In essence, the effectiveness of incorporating synthetic data for pre-training is amplified in data-scarce environments, underscoring its potential to significantly boost model performance even when real-world data is limited.

Based on these results, synthesizing ECGs using domain knowledge to train DNN models is considered highly useful for classification tasks involving rare diseases with limited data. 
Rare diseases often suffer from data scarcity, making high-accuracy classification challenging with traditional methods. 
However, by using synthetic data, it is possible to build high-performance models even with limited real-world data. 
This approach can lead to early detection and accurate diagnosis of rare diseases, thereby contributing to improved diagnostic accuracy and patient treatment outcomes in the medical field.


This study is subject to several limitations. First, the proposed simulator is designed to generate diverse ECG-like waveforms for model training and synthetic-to-real transfer, not to serve as a clinically complete physiological simulator. 
Because the waveforms are generated from simple Gaussian components and manually specified abnormality-specific rules, they may not capture all demographic, device-specific, multi-lead, or rare pathological variations present in real ECGs. 
Second, the manual development of class-specific synthesis rules entails substantial costs, especially when expanding the scope to a wider range of ECG abnormalities. 
Third, the precise relationship between the fidelity of synthesized ECGs and their downstream performance characteristics remains insufficiently understood. 
Therefore, synthesized ECGs should be viewed as a training resource for improving model robustness under data scarcity, rather than as a replacement for clinically representative validation data.

\section*{Acknowledgement}

This work was supported by JSPS KAKENHI Grant Number JP22K18001 (to NN).

%
%
\bibliographystyle{splncs04}
\bibliography{citations}
%





\newpage
\appendix
\section*{Appendix}
\setcounter{table}{0}
\renewcommand{\thetable}{A.\arabic{table}}
\setcounter{figure}{0}
\renewcommand{\thefigure}{A.\arabic{figure}}

\renewcommand{\thesection}{\Alph{section}}

\section{Details of ECG synthesis}
\label{app:ecg_synthesis}

This appendix provides additional details of the Gaussian-composition ECG simulator used to generate the synthetic ECGs in this study. The simulator generates single-lead II ECG-like signals by repeatedly composing Gaussian-shaped P, Q, R, S, and T wave components. It is intended to provide diverse synthetic training data for abnormal ECG classification, rather than to serve as a clinically complete physiological simulator.

\subsection{Synthesis procedure}

Given a target length of 5,000 time steps, corresponding to 10 seconds at 500 Hz, we generate a synthetic ECG as follows.

\begin{enumerate}
    \item Initialize an empty ECG signal.
    \item Sample the initial P, Q, R, S, and T wave parameters from Gaussian distributions using the base values and inter-sample standard deviations in Table~\ref{tab:gaussian_parameters}.
    \item Generate one heartbeat by summing the Gaussian-shaped P, Q, R, S, and T wave components.
    \item Append the generated heartbeat to the ECG signal.
    \item Perturb the waveform parameters using the beat-level standard deviations in Table~\ref{tab:gaussian_parameters}.
    \item Repeat Steps 3--5 until the signal exceeds the target length.
    \item Trim the signal to 5,000 time steps.
    \item Add white noise and sinusoidal baseline fluctuation.
    \item Standardize the sample by subtracting its mean and dividing by its standard deviation.
\end{enumerate}

For a waveform component $w$, the Gaussian peak is computed as
\[
g_w(t)
=
a_w
\exp \left(
-\frac{1}{2}
\left(
\frac{t-\mu_w}{\sigma_w}
\right)^2
\right),
\]
where $a_w$ is the signed amplitude, $\mu_w$ is the temporal shift, and $\sigma_w$ is the width of the component. By summing the P, Q, R, S, and T components, we obtain one synthetic heartbeat.

\subsection{Parameterization}

Table~\ref{tab:gaussian_parameters} shows the parameters used in the Gaussian-composition ECG simulator. The ``Base'' column denotes the mean value of each parameter. The ``Inter-sample SD'' column denotes the standard deviation used when sampling the initial parameter value for each synthetic recording. The ``Beat-level SD'' column denotes the standard deviation of the perturbation applied between consecutive beats within a recording.

\begin{table}[t]
\centering
\caption{Parameters used in the Gaussian-composition ECG simulator. Shift and width are defined in normalized beat time.}
\label{tab:gaussian_parameters}
\begin{tabular}{llccc}
\hline
Parameter & Wave & Base & Inter-sample SD & Beat-level SD \\
\hline
Amplitude & P & 0.3000 & 0.2000 & 0.1000 \\
Amplitude & Q & 0.1000 & 0.1500 & 0.0160 \\
Amplitude & R & 1.5000 & 0.7500 & 0.2500 \\
Amplitude & S & 0.5000 & 0.7000 & 0.0500 \\
Amplitude & T & 0.4500 & 0.5000 & 0.0700 \\
\hline
Width & P & 0.0300 & 0.0200 & 0.0050 \\
Width & Q & 0.0500 & 0.0500 & 0.0125 \\
Width & R & 0.0075 & 0.0100 & 0.00187 \\
Width & S & 0.0400 & 0.0400 & 0.0100 \\
Width & T & 0.0300 & 0.0300 & 0.0075 \\
\hline
Shift & P & 0.2000 & 0.1500 & 0.0200 \\
Shift & Q & 0.3000 & 0.1000 & 0.0100 \\
Shift & R & 0.3450 & 0.0200 & 0.0045 \\
Shift & S & 0.4400 & 0.0500 & 0.0150 \\
Shift & T & 0.5900 & 0.0600 & 0.0150 \\
\hline
\end{tabular}
\end{table}

The base values were manually selected so that a generated beat exhibits a typical lead-II P-QRS-T morphology. The inter-sample perturbations were set larger than the beat-level perturbations to increase diversity across synthetic recordings while maintaining temporal consistency within each recording. Because these parameters are intentionally broad, some generated waveforms may deviate from typical physiological ECG morphology. We therefore treat the synthesized signals as a training and augmentation resource, not as a substitute for clinically representative ECG data.
\section{Classification performance evaluated with AUPRC}
\label{app:auprc}

\begin{table}[h!]
\centering
\caption{
AFLT classification performance (AUPRC) 
}\label{tab01a:aflt}
    \begin{threeparttable}
    \begin{tabularx}{\textwidth}{
        X C C C D
    }
    \toprule
     & $a.$ Real & $b.$ Syn & $c.$ Syn $\rightarrow$ Real & Gain (\%)\tnote{*}\\
    \midrule
    $n$ synthesized\tnote{\dag} 
        & \multicolumn{1}{r}{$0\quad\quad$} 
        & \multicolumn{1}{r}{$10,000\quad\quad$} 
        & \multicolumn{1}{r}{$10,000\quad\quad$}  & - \\
    $n$ real-world\tnote{\dag} 
        & \multicolumn{1}{r}{$4,633\quad\quad$} 
        & \multicolumn{1}{r}{$0\quad\quad$} 
        & \multicolumn{1}{r}{$4,633\quad\quad$}  & - \\
    \midrule
    EfficientNet-B0
	&$0.9967\pm0.0067$
	&$0.0834\pm0.0715$
	&$1.0000\pm0.0000$
	&$0.3311$ \\
    GRU
	&$0.9745\pm0.0289$
	&$0.1302\pm0.0661$
	&$0.9967\pm0.0067$
	&$2.2781$ \\
    LSTM
	&$0.9987\pm0.0026$
	&$0.0527\pm0.0197$
	&$1.0000\pm0.0000$
	&$0.1302$\\
    Luna
	&$0.9550\pm0.0740$
	&$0.1870\pm0.1026$
	&$1.0000\pm0.0000$
	&$4.712$ \\
    Mega
	&$0.9756\pm0.0195$
	&$0.0854\pm0.0830$
	&$0.9987\pm0.0026$
	&$2.3678$\\
    ResNet18
	&$0.9928\pm0.0114$
	&$0.0879\pm0.0778$
	&$1.0000\pm0.0000$
	&$0.7252$\\
    ResNet34
	&$0.8948\pm0.1757$
	&$0.2033\pm0.2883$
	&$1.0000\pm0.0000$
	&$11.7568$ \\
    ResNet50
	&$0.9660\pm0.0483$
	&$0.1011\pm0.0722$
	&$0.9938\pm0.0078$
	&$2.8778$\\
    S4	
	&$0.9452\pm0.1064$
	&$0.0681\pm0.0344$
	&$0.9918\pm0.0104$
	&$4.9302$\\
    Transformer
	&$0.8988\pm0.1314$
	&$0.0532\pm0.0475$
	&$1.0000\pm0.0000$
	&$11.2595$ \\
    \midrule
    Average
	&$0.9598$
	&$0.1052$
	&$0.9981$
	&$4.1369$ \\ 
    \bottomrule
    \end{tabularx}
  \begin{tablenotes}
    \small
    \item[*] The relative improvement from ``Real'' to ``Syn$\rightarrow$Real'', calculated as $(c - a) / a \times 100$.
    \item[\dag] The total number of synthesized and real-world data used for training the model, respectively.
    \end{tablenotes}
\end{threeparttable}
\end{table}

The results of classifying AFLT state ECG from normal ECG evaluated with AUPRC are presented in Table~\ref{tab01a:aflt}.
For the model trained under ``Real'' setting, the EfficientNet-B0 achieved the best performance, with a mean AUPRC of $0.9967$.
In contrast, when trained under ``Syn'' setting, the ResNet34 outperformed others, albeit with a significantly lower mean AUPRC of $0.2033$.
Interestingly, with ``Syn $\rightarrow$ Real'' setting, the EfficientNet-B0, LSTM, Luna, ResNet18, ResNet34, and Transformer architecture achieved the best performance, with an AUPRC of $1.0000$.
On average, across the ten models, the ``Syn $\rightarrow$ Real'' approach demonstrated a $4.14\%$ improvement over the models trained exclusively with real-world data.

\begin{table}[h!]
\centering
\caption{
WPW classification performance (AUPRC)
}\label{tab01a:wpw}
    \begin{threeparttable}
    \begin{tabularx}{\textwidth}{
        X C C C D
    }
    \toprule
     & $a.$ Real & $b.$ Syn & $c.$ Syn $\rightarrow$ Real & Gain (\%)\tnote{*}\\
    \midrule
    $n$ synthesized\tnote{\dag} 
        & \multicolumn{1}{r}{$0\quad\quad$} 
        & \multicolumn{1}{r}{$10,000\quad\quad$} 
        & \multicolumn{1}{r}{$10,000\quad\quad$}  & - \\
    $n$ real-world\tnote{\dag} 
        & \multicolumn{1}{r}{$4,642\quad\quad$} 
        & \multicolumn{1}{r}{$0\quad\quad$} 
        & \multicolumn{1}{r}{$4,642\quad\quad$}  & - \\
    \midrule
    EfficientNet-B0
	&$0.6174\pm0.3068$
	&$0.0421\pm0.0273$
	&$0.7777\pm0.1516$
	&$25.9637$ \\
    GRU
	&$0.7060\pm0.0347$
	&$0.0704\pm0.0219$
	&$0.7178\pm0.0684$
	&$1.6714$ \\
    LSTM
	&$0.7716\pm0.0326$
	&$0.0841\pm0.0499$
	&$0.6313\pm0.1149$
	&$-18.183$\\
    Luna
	&$0.3469\pm0.1665$
	&$0.0130\pm0.0028$
	&$0.1936\pm0.2168$
	&$-44.1914$ \\
    Mega
	&$0.6630\pm0.1131$
	&$0.0307\pm0.0033$
	&$0.6907\pm0.0694$
	&$4.178$ \\
    ResNet18
	&$0.5476\pm0.2487$
	&$0.0366\pm0.0182$
	&$0.7793\pm0.1086$
	&$42.3119$ \\
    ResNet34
	&$0.5745\pm0.1931$
	&$0.0728\pm0.0154$
	&$0.7748\pm0.0696$
	&$34.8651$ \\
    ResNet50
	&$0.3213\pm0.2872$
	&$0.0197\pm0.0085$
	&$0.6769\pm0.1032$
	&$110.6754$ \\
    S4	
	&$0.6523\pm0.1184$
	&$0.0452\pm0.0138$
	&$0.4652\pm0.1417$
	&$-28.6831$ \\
    Transformer
	&$0.5678\pm0.3093$
	&$0.0792\pm0.1123$
	&$0.6789\pm0.0649$
	&$19.5667$ \\
    \midrule
    Average
	&$0.5768$
	&$0.0494$
	&$0.6386$
	&$14.8175$ \\    
    \bottomrule
    \end{tabularx}
    \begin{tablenotes}
    \small
    \item[*] The relative improvement from ``Real'' to ``Syn$\rightarrow$Real'', calculated as $(c - a) / a \times 100$.
    \item[\dag] The total number of synthesized and real-world data used for training the model, respectively.
    \end{tablenotes}
\end{threeparttable}
\end{table}

The results of classifying WPW state ECG from normal ECG evaluated with AUPRC are presented in Table~\ref{tab01a:wpw}.
For the model trained under ``Real'' setting, the LSTM achieved the best performance, with a mean AUPRC of $0.7716$.
In contrast, when trained under ``Syn'' setting, the LSTM outperformed others, albeit with a significantly lower mean AUPRC of $0.0841$.
Interestingly, with ``Syn $\rightarrow$ Real'' setting, the ResNet18 architecture achieved the best performance, with an AUPRC of $0.7793$.
On average, across the ten models, the ``Syn $\rightarrow$ Real'' approach demonstrated a $14.82\%$ improvement over the models trained exclusively with real-world data.

\begin{table}[h!]
\centering
\caption{
PVC classification performance (AUPRC) 
}\label{tab01a:pvc}
    \begin{threeparttable}
    \begin{tabularx}{\textwidth}{
        X C C C D
    }
    \toprule
     & $a.$ Real & $b.$ Syn & $c.$ Syn $\rightarrow$ Real & Gain (\%)\tnote{*}\\
    \midrule
    $n$ synthesized\tnote{\dag} 
        & \multicolumn{1}{r}{$0\quad\quad$} 
        & \multicolumn{1}{r}{$10,000\quad\quad$} 
        & \multicolumn{1}{r}{$10,000\quad\quad$}  & - \\
    $n$ real-world\tnote{\dag} 
        & \multicolumn{1}{r}{$5,257\quad\quad$} 
        & \multicolumn{1}{r}{$0\quad\quad$} 
        & \multicolumn{1}{r}{$5,257\quad\quad$}  & - \\
    \midrule
    EfficientNet-B0	
	&$0.9660\pm0.0051$
	&$0.5680\pm0.0972$
	&$0.9681\pm0.0052$
	&$0.2174$ \\    
    GRU
	&$0.9710\pm0.0064$
	&$0.5263\pm0.0546$
	&$0.9629\pm0.0079$
	&$-0.8342$\\    
    LSTM
	&$0.9668\pm0.0035$
	&$0.5520\pm0.0700$
	&$0.9692\pm0.0048$
	&$0.2482$\\    
    Luna
	&$0.8840\pm0.0258$
	&$0.5020\pm0.1848$
	&$0.9018\pm0.0051$
	&$2.0136$ \\    
    Mega
	&$0.9676\pm0.0074$
	&$0.5523\pm0.0868$
	&$0.9642\pm0.0111$
	&$-0.3514$\\    
    ResNet18
	&$0.9580\pm0.0088$
	&$0.3085\pm0.0197$
	&$0.9606\pm0.0090$
	&$0.2714$\\    
    ResNet34
	&$0.9525\pm0.0070$
	&$0.3170\pm0.0994$
	&$0.9697\pm0.0045$
	&$1.8058$\\    
    ResNet50	
	&$0.9336\pm0.0072$
	&$0.3077\pm0.0823$
	&$0.9534\pm0.0055$
	&$2.1208$\\    
    S4
	&$0.9603\pm0.0072$
	&$0.5590\pm0.0329$
	&$0.9601\pm0.0084$
	&$-0.0208$ \\  
    Transformer
	&$0.8960\pm0.0105$
	&$0.5817\pm0.0290$
	&$0.9347\pm0.0072$
	&$4.3192$ \\    
    \midrule
    Average
	&$0.9456$
	&$0.4775$
	&$0.9545$
	&$0.9790$ \\    
    \bottomrule
    \end{tabularx}
  \begin{tablenotes}
    \small
    \item[*] The relative improvement from ``Real'' to ``Syn$\rightarrow$Real'', calculated as $(c - a) / a \times 100$.
    \item[\dag] The total number of synthesized and real-world data used for training the model, respectively.
    \end{tablenotes}
\end{threeparttable}
\end{table}

The results of classifying PVC state ECG from normal ECG evaluated with AUPRC are presented in Table~\ref{tab01a:pvc}.
For the model trained under ``Real'' setting, the Mega achieved the best performance, with a mean AUPRC of $0.9676$.
In contrast, when trained under ``Syn'' setting, the Transformer outperformed others, albeit with a significantly lower mean AUPRC of $0.5817$.
Interestingly, with ``Syn $\rightarrow$ Real'' setting, the ResNet34 architecture achieved the best performance, with an AUPRC of $0.9697$.
On average, across the ten models, the ``Syn $\rightarrow$ Real'' approach demonstrated a $0.979\%$ improvement over the models trained exclusively with real-world data.

\begin{table}[h!]
\centering
\caption{
AF classification performance (AUPRC)
}\label{tab01a:af}
    \begin{threeparttable}
    \begin{tabularx}{\textwidth}{
        X C C C D
    }
    \toprule
     & $a.$ Real & $b.$ Syn & $c.$ Syn $\rightarrow$ Real & Gain (\%)\tnote{*}\\
    \midrule
    $n$ synthesized\tnote{\dag} 
        & \multicolumn{1}{r}{$0\quad\quad$} 
        & \multicolumn{1}{r}{$10,000\quad\quad$} 
        & \multicolumn{1}{r}{$10,000\quad\quad$}  & - \\
    $n$ real-world\tnote{\dag} 
        & \multicolumn{1}{r}{$5,566\quad\quad$} 
        & \multicolumn{1}{r}{$0\quad\quad$} 
        & \multicolumn{1}{r}{$5,566\quad\quad$}  & - \\
    \midrule
    EfficientNet-B0	
	&$0.9871\pm0.0053$
	&$0.3899\pm0.0671$
	&$0.9879\pm0.0022$
	&$0.081$\\    
    GRU
	&$0.9881\pm0.0061$
	&$0.7938\pm0.0169$
	&$0.9910\pm0.0016$
	&$0.2935$\\    
    LSTM
	&$0.9885\pm0.0048$
	&$0.5890\pm0.3559$
	&$0.9811\pm0.0108$
	&$-0.7486$ \\    
    Luna
	&$0.9807\pm0.0034$
	&$0.5201\pm0.0280$
	&$0.9739\pm0.0021$
	&$-0.6934$\\    
    Mega
	&$0.9805\pm0.0030$
	&$0.7258\pm0.0435$
	&$0.8255\pm0.3257$
	&$-15.8083$\\    
    ResNet18
	&$0.9904\pm0.0005$
	&$0.3400\pm0.0557$
	&$0.9897\pm0.0017$
	&$-0.0707$\\    
    ResNet34
	&$0.9882\pm0.0026$
	&$0.3610\pm0.0553$
	&$0.9907\pm0.0020$
	&$0.253$ \\    
    ResNet50
	&$0.9893\pm0.0023$
	&$0.4292\pm0.0840$
	&$0.9898\pm0.0019$
	&$0.0505$ \\    
    S4
	&$0.9890\pm0.0008$
	&$0.6749\pm0.0204$
	&$0.9899\pm0.0009$
	&$0.091$ \\    
    Transformer
	&$0.9806\pm0.0033$
	&$0.6254\pm0.0479$
	&$0.9723\pm0.0036$
	&$-0.8464$ \\    
    \midrule
    Average 
	&$0.9862$
	&$0.5449$
	&$0.9692$
	&$-1.7398$ \\    
    \bottomrule
    \end{tabularx}
  \begin{tablenotes}
    \small
    \item[*] The relative improvement from ``Real'' to ``Syn$\rightarrow$Real'', calculated as $(c - a) / a \times 100$.
    \item[\dag] The total number of synthesized and real-world data used for training the model, respectively.    
    \end{tablenotes}
\end{threeparttable}
\end{table}

The results of classifying AF state ECG from normal ECG evaluated with AUPRC are presented in Table~\ref{tab01a:af}.
For the model trained under ``Real'' setting,  the ResNet18 achieved the best performance, with a mean AUPRC of $0.9904$.
In contrast, when trained under ``Syn'' setting, the GRU outperformed others, albeit with a significantly lower mean AUPRC of $0.7938$.
With ``Syn $\rightarrow$ Real'' setting, the GRU architecture achieved the best performance, with an AUPRC of $0.9910$.
On average, across the ten models, the ``Syn $\rightarrow$ Real'' approach demonstrated a $1.74\%$ deterioration over the models trained exclusively with real-world data.

\end{document}